\documentclass[letterpaper, 10 pt, conference]{ieeeconf}  

\IEEEoverridecommandlockouts                              

\usepackage[T1]{fontenc}
\usepackage{graphicx}
\usepackage{subcaption}
\usepackage{xcolor}
\usepackage[]{footmisc}
\usepackage{stfloats}

\usepackage{algorithm}
\usepackage{algorithmic}
\usepackage{subcaption}
\usepackage{tabularx,booktabs}
\usepackage{graphicx}
\usepackage{float}
\usepackage{amsmath}

\usepackage{balance}

\usepackage[backend=biber,style=numeric,sorting=none, style=ieee]{biblatex}
\usepackage{multirow}

\title{\LARGE \bf
Increasing Transparency of Reinforcement Learning using Shielding for Human Preferences and Explanations
}

\author{Georgios Angelopoulos$^{1}$, Luigi Mangiacapra$^{2}$, Alessandra Rossi$^{2}$, Claudia {Di Napoli}$^{3}$, and Silvia Rossi$^{2}$
\thanks{This work has been supported by the European Union's Horizon 2020 research and innovation programme under the Marie Sk\l{}odowska-Curie grant agreement No 955778.}
\thanks{$^{1}$Georgios Angelopoulos is with the Interdepartmental Center for Advances in Robotic Surgery - ICAROS, University of Naples Federico II, Naples, Italy {\tt\small georgios.angelopoulos@unina.it}}%
\thanks{$^{2}$Luigi Mangiacapra, Alessandra Rossi and Silvia Rossi are with the Department of Electrical Engineering and Information Technologies - DIETI, University of Naples Federico II, Napoli, Italy {\tt\small \{alessandra.rossi, silvia.rossi\}@unina.it}}%
\thanks{$^{3}$Claudia Di Napoli is with the Institute for High Performance Computing and Networking, CNR, Naples, Italy {\tt\small claudia.dinapoli@cnr.it}}%
}

\begin{document}

\maketitle
\thispagestyle{empty}
\pagestyle{empty}

\begin{abstract}

The adoption of Reinforcement Learning (RL) in several human-centred applications provides robots with autonomous decision-making capabilities and adaptability based on the observations of the operating environment. In such scenarios, however, the learning process can make robots' behaviours unclear and unpredictable to humans, thus preventing a smooth and effective Human-Robot Interaction (HRI). As a consequence, it becomes crucial to avoid robots performing actions that are unclear to the user. In this work, we investigate whether including human preferences in RL (concerning the actions the robot performs during learning) improves the transparency of a robot's behaviours. For this purpose, a shielding mechanism is included in the RL algorithm to include human preferences and to monitor the learning agent's decisions. We carried out a within-subjects study involving 26 participants to evaluate the robot's transparency in terms of Legibility, Predictability, and Expectability in different settings. Results indicate that considering human preferences during learning improves Legibility with respect to providing only Explanations, and combining human preferences with explanations elucidating the rationale behind the robot's decisions further amplifies transparency. Results also confirm that an increase in transparency leads to an increase in the safety, comfort, and reliability of the robot. These findings show the importance of transparency during learning and suggest a paradigm for robotic applications with human in the loop.

\end{abstract}
\section{Introduction}

Non-industrial settings require robots to autonomously acquire new skills, as it is impractical to pre-program all potential knowledge they might need to collaborate with people. In order to allow users to confidently decide when and how to collaborate with the robot, knowing its current skill level and capabilities, they also need to be able to assess the robot's proficiency level in a specific task. It becomes, therefore, essential for robots to exhibit transparent behaviours while learning.

In close HRI scenarios \cite{kirschner2022expectable}, the behaviours of robots must be transparent for effective interaction. Indeed, when the human is involved in the learning process, effective robot learners need to behave in a transparent way \cite{breazeal2009role}, since the lack of transparency may result in the human seeing the robot as a black box system, but often conventional methods to achieve transparency may lead to overly conservative policies or the freezing robot problem \cite{trautman2010unfreezing}.
In \cite{wang2022transparency}, the authors discussed that the transparency of a system (in our case, a robot) could potentially be increased through personalisation. A personalised robot should take into account human preferences as explained in \cite{syrdal2007personalized}.

Moreover, standard learning mechanisms in robots do not assure that human-adapted and desirable decisions are taken into account while learning. For example, during the exploration phase in Reinforcement Learning, robots may choose actions leading to undesirable outcomes that may contrast with the user expectations. While various approaches \cite{arakawa2018dqn, hadfield2017inverse, de2019accelerating} have been proposed to address this, such as encoding negative rewards with handcrafting policies or limiting the set of applicable actions in each Markov Decision Process state, they still proved to be ineffective \cite{wei2022incorporating}.

In this direction, this work proposes an RL mechanism that includes user preferences to increase the transparency of the robot's behaviour.
As reported in \cite{endsley2017here, alonso2018system}, transparency is influenced by Legibility, Expectability, and Predictability factors. Furthermore, a recent study in the field \cite{schott2023literature} has underscored the pivotal role of explainability in transparency. Therefore, we also integrate explanations of the robot's actions when preferences are considered, as these explanations help users understand why the robot is making certain decisions, making its behaviour more predictable and its intentions more legible to users. 

The proposed mechanism is inspired by safe Reinforcement Learning techniques, but instead of introducing constraints during learning, preferences are introduced to allow the robot to learn personalised policies and, consequently, increase the robot's transparency. We propose to use a shielding mechanism \cite{alshiekh2018safe} for RL to include human preferences as, for example, a preferred direction for a navigation task.
Shielding is a control mechanism that we use to account for human preferences by ranking the available actions of the learning agents. It supervises the decision-making process to replace a selected action with a user's preferable one if available and applicable \cite{alshiekh2018safe}.

The current study focuses on designing and testing a learning mechanism that is comprehensible to humans by integrating human preferences and providing explanations of actions through a shielding mechanism. 
The objective of this study is to evaluate the transparency of these mechanisms and assess their impact on the social attribution of the robot. The results of this research aim to contribute to the design of socially acceptable robots that can coexist harmoniously with humans by increasing the transparency of their behaviour.

\section{Related Work}

Several approaches have been proposed in the literature to include human preferences and constraints during robot learning. However, how to facilitate comfortable and socially acceptable robot behaviour while learning is still relatively unexplored, and it needs more attention. 

One approach to include human preferences in the learning process is to include human-in-the-loop during the learning, as it is proposed in Interactive Reinforcement Learning (IntRL). IntRL solves decision-making tasks by incorporating human feedback as numeric rewards, action corrections, and preferences. For example, Tsiakas et al. \cite{tsiakas2018task} proposed an IntRL framework that uses human feedback to achieve an efficient personalisation that respects human constraints in a cognitive task. Recent works \cite{10.1007/978-3-031-24667-8_27, matarese2021toward} proposed a model to improve robot transparency during IntRL by incorporating non-verbal emotional and behavioural responses into a humanoid robot. These works emphasise the importance of transparency in RL and the potential of human feedback, but they require constant human advice during learning. 

Another approach is Inverse Reinforcement Learning (IRL), where the reward function is inferred by observing the human behaviour in performing the task. The observed behaviour can provide insight into the implicit constraints or preferences essential to the task. Andriella et al. \cite{andriella2022introducing} presented a novel framework utilising IRL that actively learns robotic assistive behaviours by leveraging the therapist's expertise and their demonstrations. Another work from Chen et al. \cite{chen2022android} showed a study using a robot trained by IRL to act as a receptionist; their results showed that the robot successfully managed to receive the knowledge and the constraints of the human. However, most IRL algorithms lack practical safety assessments; they are often considered a black box to humans, require numerous demonstrations, and have high computational costs \cite{brown2020safe}. 

In safe Reinforcement Learning, constraints are integrated into the learning process both to guarantee that unsafe actions are not performed and to enforce a given policy. Van Waveren et al. \cite{van2022correct} proposed a shielding mechanism to limit unsafe robot actions. They tested their approach in simulated kitchen scenarios with a simulated robot, and their results showed that the shields were robust in avoiding undesired outcomes. Marta et al. \cite{marta2021human} presented a shield synthesis framework to obtain RL policies that humans perceive as safe in a Lunar landing RL benchmark and a scenario where a simulated robot navigates around humans. Their results converged into desirable policies for humans. 
However, despite the promising results of these studies, they were performed in simulated environments and did not evaluate the human perception of the robots.

In recent work, Brawer et al. \cite{brawer2023interactive} proposed the Transparent Matrix Overlay system, a novel interactive policy-shaping approach that can extract and utilize symbolic rules from user-provided directives. The overlay rules act as mutable and composable high-level constraints on the robot’s policy, allowing a user to quickly influence it in a naturalistic way without permanently altering it. Their interesting approach results in fewer user-provided corrections to the robot’s behaviour. However, akin to the research conducted by Van Waveren et al., an aspect that warrants further exploration is the evaluation of how human users perceive these systems.

Considering the mentioned studies and their limitations, this work aims to present a step forward in developing a transparent learning framework to let humans understand robots during the learning process.

\section{The Proposed Approach}

Building on the knowledge that both explanations \cite{felzmann2019robots} and personalisation \cite{wang2022transparency} of the system can improve the transparency of the system,  
our current research investigates how the integration of human preferences and explanations influences the transparency of a robot learning through Reinforcement Learning. Reinforcement Learning is a widely adopted technique to enable robots to learn optimal policies for achieving specific goals. The RL problem can be formalised as a Markov Decision Process (MDP), which is a 4-tuple representation of the environment $\{s, a, p, r \}$, including states, actions, state transition probability, and rewards. More specifically: 
\begin{itemize}
\item $\mathcal{S}$ represents a finite set of states,

\item $\mathcal{A}$ represents a set of actions that the robot can perform,

\item $P$ is a state transition function $P: \mathcal{S} \times \mathcal{A} \rightarrow \Pi(\mathcal{S})$, where $\Pi(\mathcal{S})$ is a probability distribution over $\mathcal{S}$ and $p(s, a, s\prime)$ is the probability of transitioning from state $s$ to $s\prime$ by performing action $a$,

\item $R: \mathcal{S} \times \mathcal{A} \rightarrow R$ is a scalar reward function.
\end{itemize}

 \begin{figure}
  \centering
  \includegraphics[width=0.5\textwidth]{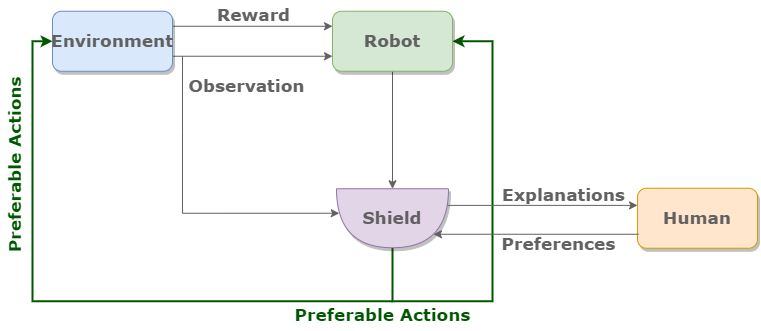}
  \caption{The proposed architecture of the learning system.}
    \label{rl}
\end{figure}

The robot's goal in RL is to learn an optimal policy. The Q-learning algorithm is commonly used to compute the optimal policy based on a state-action value function $Q(s, a)$. The algorithm updates the Q value for a given state-action pair based on the immediate reward and the maximum Q value for the next state $Q(s,a) = Q(s,a) + \alpha\cdot(r(s) + \gamma \cdot \max_{a\prime}Q( s\prime,a\prime) - Q (s,a))$, where $Q(s, a)$ is the state-action value, $\alpha$ is the learning rate in the range of 0 to 1, $r(s, a)$ is the direct reward value for the state-action pair, and $\gamma$ is the discount factor.

To integrate user preferences and explanations during learning, we propose to include a control mechanism called \textit{shield}, which, in addition to the knowledge about unsafe states, it also has knowledge about user-preferred actions. As depicted in Figure \ref{rl}, at each step of the reinforcement learning process, the learning agent selects an action as $\alpha^i_t$, which is sent to the shield for evaluation. 

The shield begins by computing a set of all actions leading to safe states, i.e. the set of safe actions for the current state $s$, $S_{safe}=\{a^1_{s}, \ldots, a^k_{s}\}$. This set is derived from the complete set of available actions, by removing those actions that would violate safety leading to an unsafe state.
If $a^i_{t} \in \{a^1_{s}, \ldots, a^k_{s}\}$, the computation proceeds. Otherwise, the shield will randomly select one $a^j_{t} \in \{a^1_{s}, \ldots, a^k_{s}\}$. This selection ensures that the chosen action is safe and respects the safety specifications.

In parallel, the shield estimates the preferred action for the current state ($a^p_s$) from the user preferences. Note that there is a single preferred action for each state. Suppose the preferred action has a Q-value greater than or equal to the one chosen by the agent and it does not lead to an unsafe state. In that case, it will replace the agent's action and will be selected for execution, since it is aligned with both safety and user preferences.

In this way, the shield considers both human preferences and the feasibility of the actions to avoid over-constraining the learning process with human preferences. The learning agent executes the action the shield selects ($a_t$), and the new state and reward information ($s\prime$ and $r\prime$) are calculated. 
The shielding mechanism is outlined in the Algorithm \ref{algori}.

It is important to note that the shield remains active even after the exploration phase is complete, as there is no guarantee that preferred actions will not be incorporated into the final policy. In this way, it is more likely that the robot, during the exploitation phase, acts toward a preferable and feasible policy. 

Finally, every time the shield chooses an alternative action, an explanation is given to the user to improve transparency. Explanations are given either when a more preferable action is chosen or when a preferable action can not be chosen. The explanation of the robots is simple and structured in a contrastive manner \cite{9562003}, such as ``the robot selected $\alpha^x$ because if it had selected $\alpha^y$ ...''. According to \cite{lyons2023explanations}, these simple explanations based on environmental observations are credible and effective types of explanations.

\begin{algorithm}
\caption{The Preference Shielding Mechanism}
\label{algori}
\begin{algorithmic}[1]
\WHILE{$s \notin \mathcal{S}{terminal}$}
\STATE Receive the current chosen action $a^i_{t}$
\STATE Compute set $S_{safe} = (a^1_s, \cdots, a^k_s)$
\IF{$a^i_{t} \notin \{a^1_{s}, \ldots, a^k_{s}\}$}
    \STATE $a_t \leftarrow rand(S_{safe})$
\ELSE
    \STATE $a_t \leftarrow a^i_{t}$
\ENDIF    
\IF{$a^p_{s} \in S_{safe}$ and $Q(s,a^p_{s}) \geq Q(s,a_t)$}
        \STATE $\alpha_t \leftarrow a^p_{s}$
\ENDIF
\IF{$\alpha_t \neq \alpha^i_t$}
    \STATE explain\_contrastive($\alpha_t,\alpha^i_t$)
\ENDIF
\ENDWHILE
\end{algorithmic}
\end{algorithm}

\section{Learning Scenario}

To test the proposed learning architecture, we employed a simple interactive scenario based on the Gridworld game. This scenario involves a grid of cells, with each cell representing a state. The robot can move between adjacent cells using predefined actions (up, down, left, right), and each state in the grid is associated with a reward. The robot's objective is to learn a navigation policy to reach a designated goal state.

\begin{figure}[t!]
  \centering
  \includegraphics[width=0.8\linewidth]{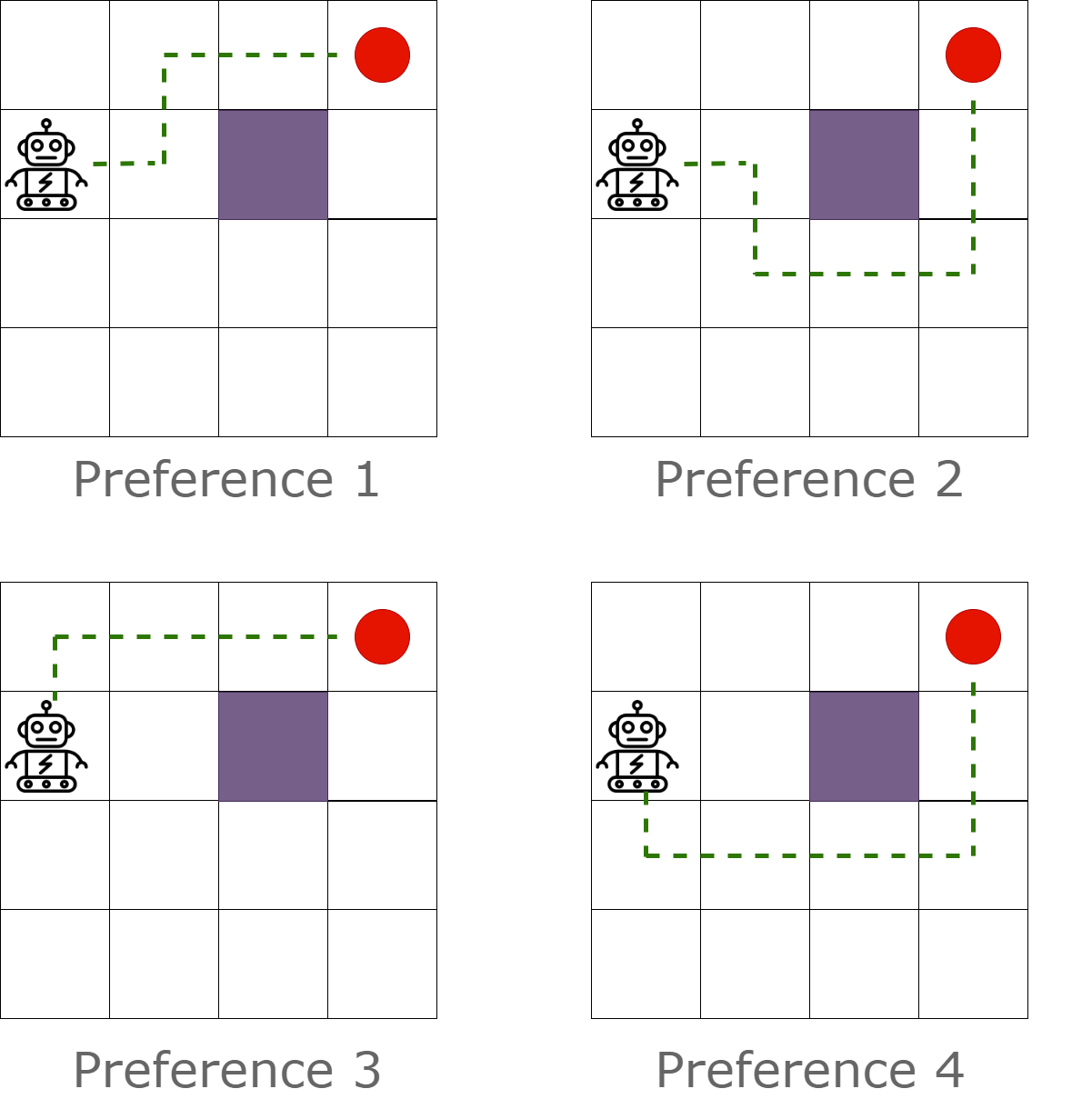}
  \caption{The four possible human preferences.}
   \label{preferences}
\end{figure}

To introduce the interaction with the human in the scenario, the participant adds obstacles to the Gridworld environment and selects the goal state. Additionally, the human provides preferences for the robot's navigation strategy in natural language. Specifically, the participant can select one preference among the following four options (see Figure \ref{preferences}): Preference 1) to avoid the obstacles in a clockwise direction; Preference 2) to avoid the obstacles in an anti-clockwise direction; Preference 3) to navigate towards the north of the map; Preference 4) to navigate towards the south of the map. 

To investigate the impact of user preferences and explanations on transparency during learning based on current literature \cite{felzmann2019robots,wang2022transparency}, we derived the following hypotheses:

\begin{itemize}
\item \textbf{\textit{Hypothesis 1} (H1)}: The learning mechanism where the robot provides explanations only produces a more transparent robot behaviour than Traditional Reinforcement Learning. 
\item \textbf{\textit{Hypothesis 2} (H2)}: The learning mechanism in which the robot incorporates human preferences into the learning process results in behaviour that is more transparent than the other learning mechanisms that do not consider human preferences. 
\item \textbf{\textit{Hypothesis 3} (H3)}: The most transparent learning mechanism is the one where the robot incorporates human preferences into the learning process and provides explanations of its actions.
\end{itemize}

These hypotheses form the basis of our investigation into the impact of human preferences and explanations on the transparency of the robot's behaviour in the Gridworld navigation scenario. For this purpose, we compare four distinct learning mechanisms employed by the robot to reach the goal within our interactive Gridworld scenario.

\begin{itemize}
\item \textbf{\textit{Learning 1} (L1)}: This mechanism integrates human preferences into the learning process (through shielding) without providing any explanations.
\item \textbf{\textit{Learning 2} (L2)}: This mechanism does not consider the human preferences given by the user, and it does not provide explanations for the robot's actions.
\item \textbf{\textit{Learning 3} (L3)}: This mechanism integrates human preferences and provides explanations for the robot's actions (through shielding).
\item \textbf{\textit{Learning 4} (L4)}: This mechanism does not consider the human preferences given by the user, but it provides explanations for the robot's actions.
\end{itemize}

\section{Experimental Method}

\begin{figure}
  \centering
  \includegraphics[height=4cm]{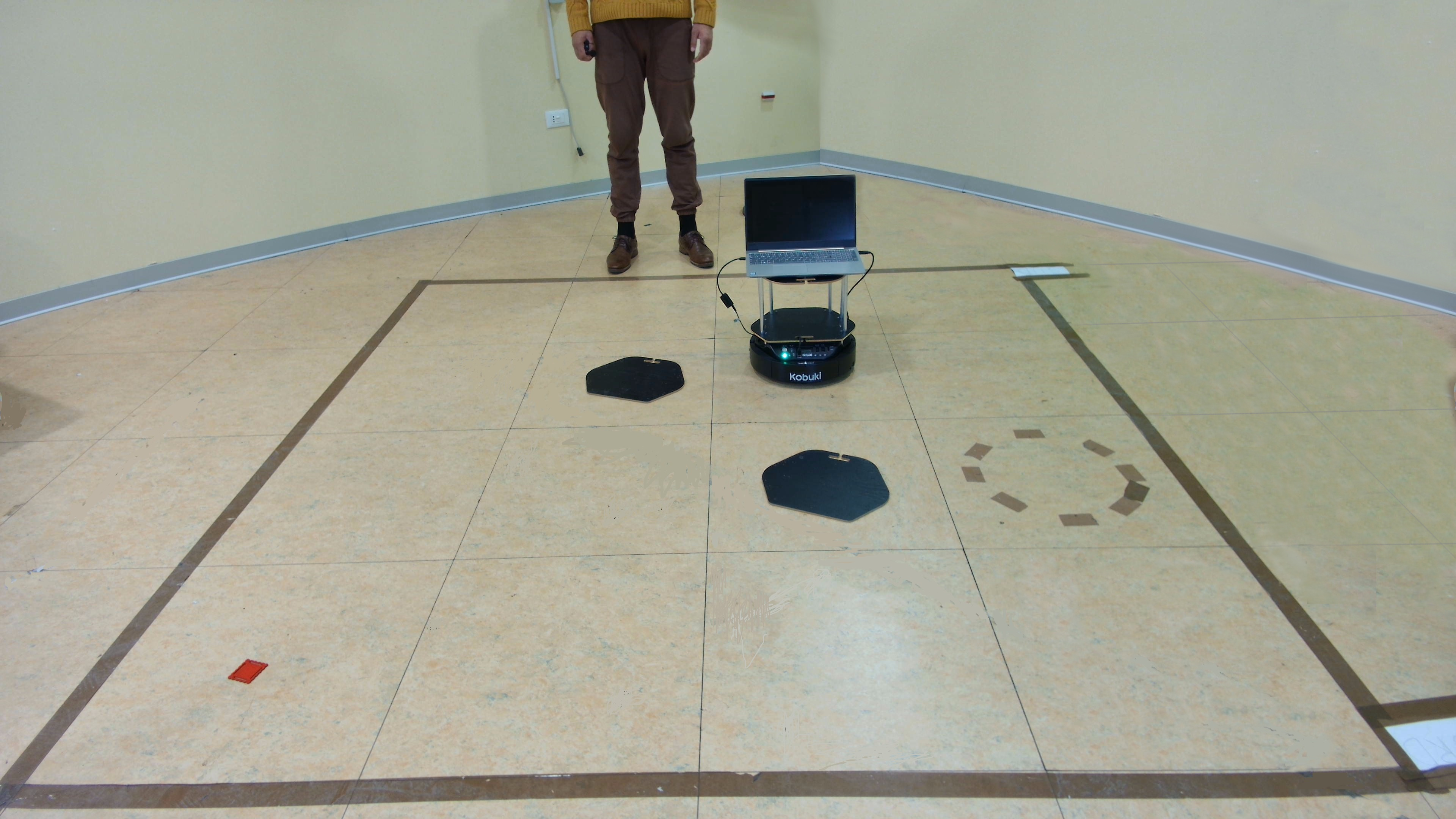}
  \caption{Before the experiment, the human gives preferences and positions obstacles and goals.}
   \label{gridworld}
\end{figure}

In this study, we aimed to evaluate the transparency of the proposed learning mechanisms through a comprehensive user study. The study was conducted at the University of Naples Federico II. It was designed as a within-subjects design, where each participant was randomly assigned the sequence of all four learning mechanisms in the designated Gridworld environment, as shown in Figure \ref{gridworld}. Notably, the institution's ethical committee granted approval for this experimentation.

\subsection{Procedure}
At the beginning of the experimental session, participants were provided with a detailed informed consent form about the research's purpose and protocol. After consenting to participate, the participants were introduced to the experimental environment and the robot. Participants were asked to place the goal placement and the objects within the designated cells in all conditions. To ensure the experiment's integrity, participants were instructed to provide the robot with a preference, even when the Learning mechanism did not account for this preference. The experimental session lasted around 20-30 minutes.

\subsection{Measurement}

To evaluate the transparency of robot behaviours, we collected participants' responses through various questionnaires. At the beginning of the experiment, we gathered demographic information about the participants (i.e., age, gender, education, and previous experience with robots), and we evaluated potential negative biases towards robots.

After the experimental trial, participants were asked to answer questions regarding Legibility, Predictability, and Expectability, which, as previously discussed, impact transparency \cite{endsley2017here, alonso2018system}: 
\begin{itemize}
    \item \textit{Legibility}: to which extent they understood why the robot moved that way;
    \item\textit{Predictability}: to which extent they understood what the robot would do next;
    \item \textit{Expectability}: to which extent the robot behaved as expected.
\end{itemize}

Then, we used a second questionnaire to measure people's sense of Safety, Comfort, and Reliability since these factors are associated with transparency \cite{lichtenthaler2014legibility}:

\begin{itemize}
    \item \textit{Safety}: to which extent they would feel safe interacting with the robot again;
    \item \textit{Comfort}: to which extent they would feel comfortable interacting with the robot again;
    \item \textit{Reliability}: to which extent they would perceive the robot reliable for future interaction.
\end{itemize}

Finally, we used the well-established and validated Human-Robot Interaction Evaluation Scale (HRIES) questionnaire \cite{spatola2021perception} to evaluate the overall people's perception of the robot in terms of Sociability, Animacy, Agency, and Disturbance and the attribution or deprivation of human characteristics. The questionnaire consisted of 16 cognitive-differences scale questions answered. Participants' responses were rated using a 7-point Likert scale (from 1 = ``Not at all'' to 7 = ``Totally'') for the three questionnaires. 

\section{Results}

We recruited 26 participants (15 male, 11 female) as predetermined a priori, aiming for an effect size of $d$=0.25 with $.85$ power at an alpha level of $.05$. The age range of participants was between 18 and 49 years ($mean=25.57$, $st.dv.=5.26$). Participants had no prior familiarity with the study setup, and 57.7\% of participants had previous exposure to robots, while 42.3\% reported having no prior interaction with robots. No negative biases towards robots were observed ($max=4$, $mean=2.19$, $st.dv.=0.93$).

\subsection{Transparency Ratings}
\begin{figure*}[th]
  \centering
  \includegraphics[width=.95\linewidth]{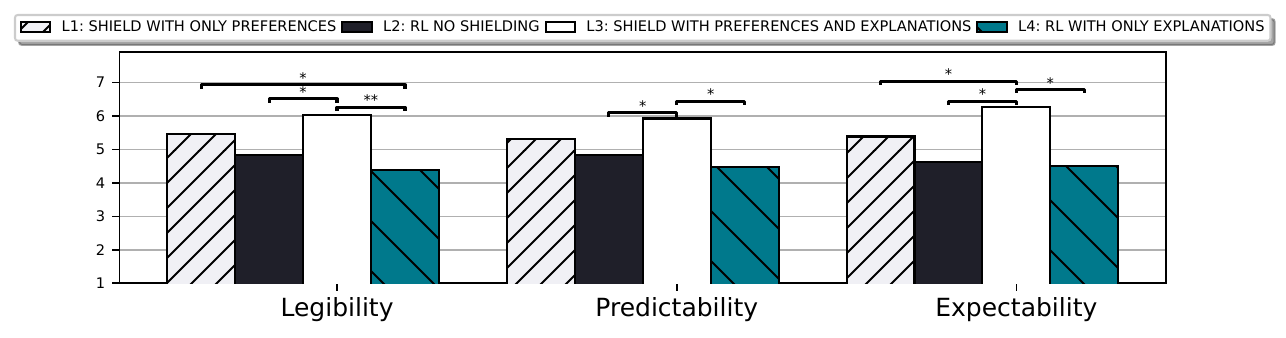}
  \caption{Legibility, Predictability, and Expectability factors for each Learning mechanism (* for $p<0.05$ and ** for $p\leq0.001$).}
  \label{transp}
\end{figure*}

A Wilcoxon signed-rank test revealed significant differences in the robot's learning mechanisms based on participants' responses, as shown in Figure \ref{transp}. The characteristic of \textit{Legibility} showed significant differences between Learning mechanisms L4 and L1 ($z$=-2.378, $p$=.017), L3 and L2 ($z$=-2.911, $p$=.004), and L4 and L3 ($z$=-3.203, $p$=.001). Similarly, \textit{Predictability} exhibited significant differences between L3 and L2 ($z$=-2.611, $p$=.009) and L4 and L3 ($z$=-3.024, $p$=.002). Moreover, \textit{Expectability} showed significant differences between L3 and L1 ($z$=-2.081, $p$=.037), L3 and L2 ($z$=-3.019, $p$=.003), and L4 and L3 ($z$=-3.050, $p$=.002).

\begin{figure*}[th]
  \centering
  \includegraphics[width=.95\linewidth]{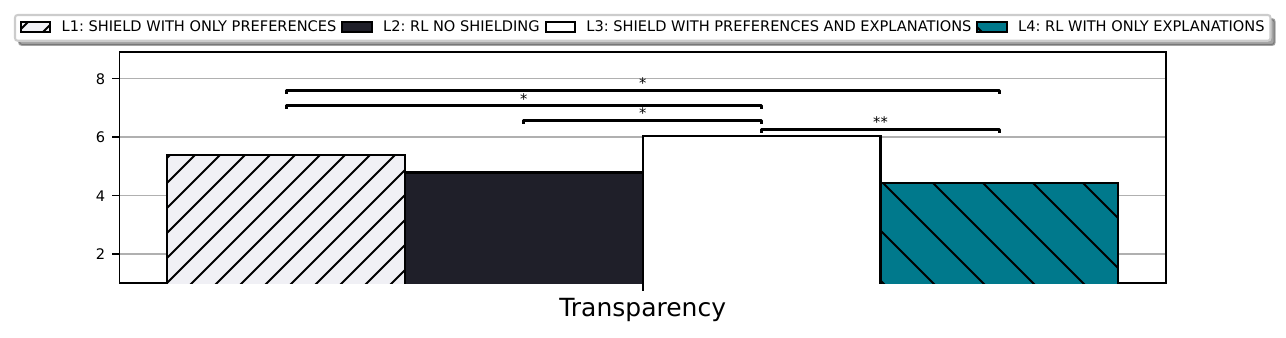}
  \caption{Transparency for each Learning mechanism (* for $p<0.05$ and ** for $p\leq0.001$).}
  \label{transpar}
\end{figure*}

It is imperative to acknowledge that the precise magnitude of influence wielded by each of the three factors (Legibility, Predictability and Expectability) on Transparency remains an open question. Therefore, to quantify Transparency, we employ the Bayesian Structural Equation Modeling (SEM) \cite{kaplan2012bayesian, girard2021reconsidering}  that combines these three key factors. Based on the output from our model, it appears that our method is acceptable. The model converged normally after 1000 iterations, and the Posterior Predictive P-value (PPP) is 0.520, indicating a good fit of the model to the data. Factor loadings obtained from the analysis reveal the strength of the relationship between each observed variable and the underlying factor (Transparency). More specifically, our data analysis found that the factor loadings for Legibility ($L$), Predictability ($P$), and Expectability ($E$) are $1.000$, $0.913$,  and $0.777$, respectively. These loadings indicate that the three components significantly contribute to Transparency, with Legibility having the highest impact, followed closely by Predictability and Expectability.

To calculate the relative importance of each component, we normalize the standardized factor loadings by dividing them by the sum of all three standardized loadings. The resulting relative importance scores are as follows: \( \mathbf{R}(\mathbf{L}):~0.385, \mathbf{R}(\mathbf{P}):~0.352, \) and \( \mathbf{R}(\mathbf{E}):~0.299 \). Utilizing these weights in a weighted sum formula, we obtain the overall transparency score (\( T \)):

\begin{equation}
T = 0.385 \cdot \text{L} + 0.352 \cdot \text{P} + 0.299 \cdot \text{E}
\end{equation}

This approach ensures that each component's contribution to the overall transparency score is appropriately weighted based on the underlying factor loadings, offering a more sophisticated measure that reflects the significance of Legibility, Predictability, and Expectability in shaping transparency. After combining the factors to compare overall transparency (see Figure \ref{transpar}), the Wilcoxon signed-rank test indicated significant differences in terms of Transparency between L4 and L3 ($z$=-3.398, $p$<.001), L3 and L2 ($z$=-3.138, $p$=.002), and L3 and L1 ($z$=-2.093, $p$=.036). A significant difference was also found between L4 and L1 ($z$=-2.175, $p$=.030).

\begin{figure*}[th]
  \centering
  \includegraphics[width=.95\linewidth]{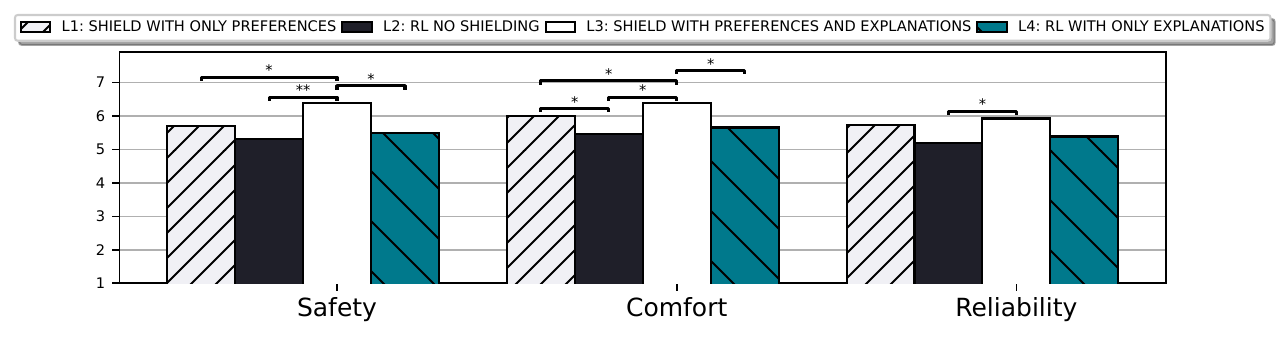}
  \caption{Safety, Comfort, and Reliability factors for each Learning mechanism (* for $p<0.05$ and ** for $p\leq0.001$).}
  \label{transp1}
\end{figure*}

We furthermore investigated the effects on safety, comfort, and reliability factors. Figure \ref{transp1} shows an increase in those factors in Learning 3. In particular, we observed  significant differences in the \textit{Safety} factor between the Learning mechanisms L3 and L1 ($z$=-2.688,  $p$=.007), L4 and L3 ($z$=-3.067,  $p$=.002) and a highly significant difference between L3 and L2 ($z$=-3.019,  $p<$.001). Regarding the \textit{Comfort} factor was also observed statistical differences between the Learning mechanisms L2 and L1 ($z$=-2.098,  $p$=.036),  L3 and L1 ($z$=-2.066,  $p$=.039), L4 and L3 ($z$=-2.809,  $p$=.005) while a highly significant difference between L3 and L2 ($z$=-3.372,  $p<$.001). Finally, a significant difference between L3 and L2 ($z$=-2.322,  $p$=.020) was observed for \textit{Reliability}.

\begin{figure*}[th]
  \centering
  \includegraphics[width=.95\linewidth]{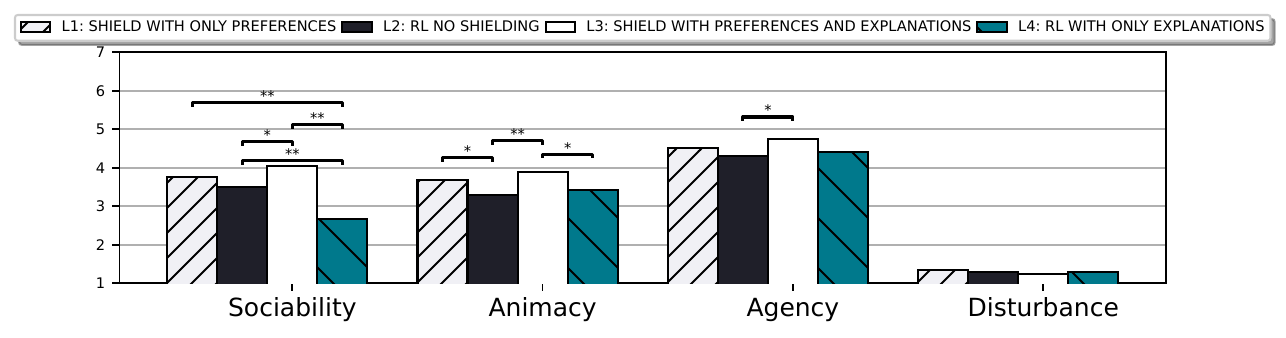}
  \caption{Average of the HRIES questionnaire responses for each Learning mechanism (* for $p<0.05$ and ** for $p\leq0.001$).}
  \label{hries}
  \vspace{-1em} 

\end{figure*}

\subsection{HRIES Ratings}

As a first step, we evaluated the internal reliability of the HRIES questionnaire for each Learning mechanism. A Cronbach's alpha test for the Sociability factor of the HRIES questionnaire was respectively $\alpha_{L1}=0.79$, $\alpha_{L2}=0.78$, $\alpha_{L3}=0.79$, $\alpha_{L4}=0.77$. The Cronbach's alpha for the Animacy factor was $\alpha_{L1}=0.80$, $\alpha_{L2}=0.77$, $\alpha_{L3}=0.73$, $\alpha_{L4}=0.76$. The Agency factor had a Cronbach's alpha of $\alpha_{L1}=0.77$, $\alpha_{L2}=0.70$, $\alpha_{L3}=0.79$, $\alpha_{L4}=0.78$. The Cronbach's alpha for the Disturbance factor was in an unacceptable range, with values of $\alpha_{L1}=0.20$, $\alpha_{L2}=0.55$, $\alpha_{L3}=0.43$, $\alpha_{L4}=0.47$. As a result, we decided to remove the two items ``Uncanny'' and ``Weird'' from the Disturbance factor. This resulted in a new Cronbach's alpha for the Disturbance factor of $\alpha_{L1}=0.77$, $\alpha_{L2}=0.78$, $\alpha_{L3}=0.75$, $\alpha_{L4}=0.74$.

A Wilcoxon signed-rank test was conducted to examine differences in the Learning mechanisms on the HRIES scale. Figure \ref{hries} shows significant differences in the factors of \textit{Sociability}, \textit{Animacy}, and \textit{Agency}, while no statistical differences were found among Learning mechanisms for the \textit{Disturbance} factor. 

In the \textit{Sociability} factor, a highly significant difference was observed between Learning 4 and Learning 1 ($z$=-3.967, $p<$.001), Learning 4 and Learning 2 ($z$=-3.325, $p<$.001), and Learning 4 and Learning 3 ($z$=-3.804, $p<$.001). We also found a significant difference between Learning 3 and Learning 2 ($z$=-2.645, $p$=.008). As a result, Learning 3 received a more positive evaluation than the other Learning mechanisms in \textit{Sociability}. This factor plays a crucial role in evaluating and accepting HRI systems. People tend to judge others based on their ability to interact positively with them \cite{spatola2021perception}.

Considering the \textit{Animacy} factor, significant differences were found between Learning 2 and Learning 1 ($z$=-2.278, $p$=.023), Learning 4 and Learning 3 ($z$=-2.018, $p$=.044) and a highly significant difference between Learning 3 and Learning 2 ($z$=-3.732, $p<$.001). Learning 3 was evaluated more positively than the other Learning mechanisms for \textit{Animacy}, suggesting that participants attributed human-like qualities to the robot when utilising Learning 3.

For the \textit{Agency} factor, a significant difference was observed between Learning 3 and Learning 2 ($z$=-3.090, $p$=.002). Sociability, Agency, and Animacy dimensions positively correlate with anthropomorphism \cite{spatola2021perception}; the results thus suggest an increase in the robot's anthropomorphism in Learning 3. This could be attributed to the increased transparency of the robotic system, as noted in recent research \cite{straten2020transparency}, which highlighted the strong impact of transparency on anthropomorphism.

\subsection{The Effect of Shielding on Learning}

To evaluate the effect of shielding on the performance of the learning mechanism, we analyse the accumulated rewards obtained with different user preferences and compare them with the accumulated rewards of the traditional RL without shielding.

\begin{figure}[t!]
  \centering
  \includegraphics[width=\linewidth]{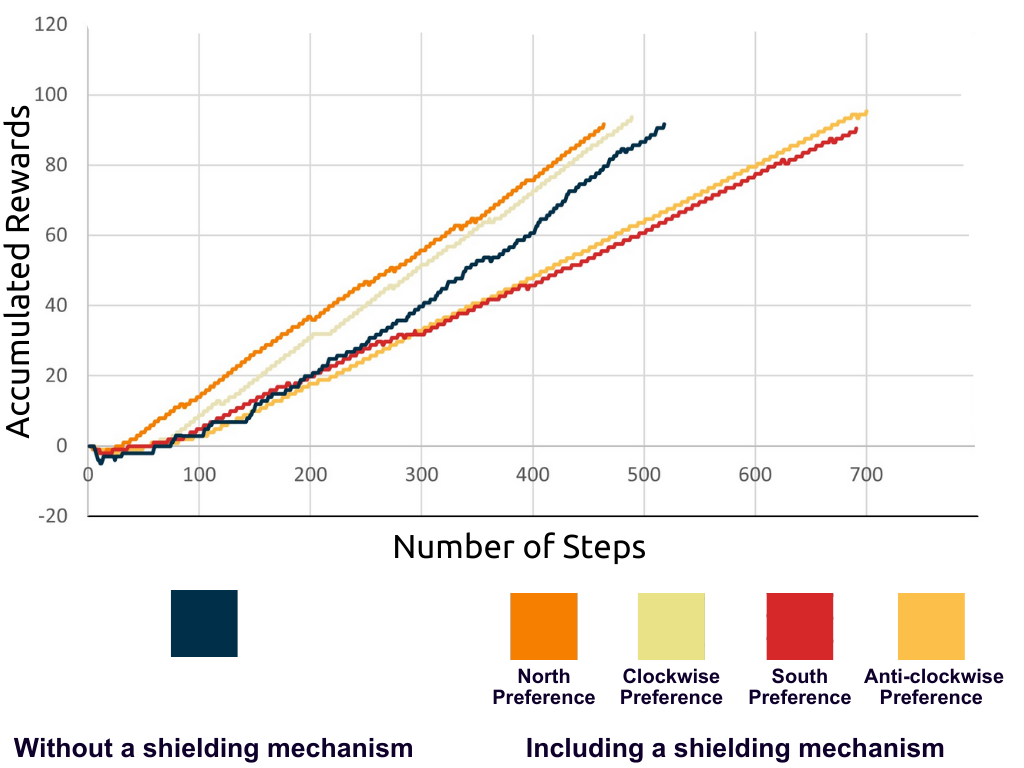}
  \caption{Accumulated Rewards.}
  \label{accum}
  \vspace{-1.7em} 

\end{figure}

As expected, when preferences align with the goal, i.e., they allow reaching the goal more quickly because they limit the number of errors during learning, the accumulated reward overcomes the one obtained without shielding. In fact, in these cases, the preferences guide the learning agent towards more rewarding actions sooner, thus leading to a faster accumulation of rewards.
As depicted in Figure \ref{accum}, this occurs when the North and Clockwise preferences are considered.
 The traditional RL shows more negative rewards at the onset, compared to the RL with shielding, since initially exploring the environment without guidance might lead to penalized actions before learning from its experiences.
Conversely, the learning mechanism with shielding incorporating human preferences introduces a control mechanism that shapes the agent's decisions.

However, the shielding mechanism with Anti-clockwise and South preferences shows a slower increase in the accumulated rewards. These preferences push the agent to choose actions or paths not directly aligned for reaching the goal, thus slowing down the learning process.
Nonetheless, when incorporating user preferences to improve transparency for an effective human-robot interaction, the learning speed is not the only factor to consider. 

In summary, incorporating human preferences via a shielding mechanism significantly impacts their learning efficiency and the rate of reward accumulation. Further, the choice between human preferences leads to the observed variations in the speed of reward accumulation.

This study aimed to evaluate the transparency of the proposed robot learning mechanisms. As shown in Figure \ref{transp}, we observed that Learning 4, which provided explanations but did not consider participants' preferences, has no statistical difference with Learning 2, which neither considered preferences nor offered explanations. Consequently, Hypothesis 1 (\textbf{H1}) was not confirmed. Additionally, participants rated the sociability of the robot in Learning 4 lower than in Learning 2, but the robot in Learning 4 exhibited higher agency than in Learning 2. This observation may be attributed to the fact that while the robot in Learning 4 provided explanations, it did not consider participants' preferences, possibly engendering a sense of disconnection or disregard for their input. This, in turn, led to reduced transparency ratings. Conversely, the robot in Learning 2, which lacked both explanations and consideration of preferences, may have set more consistent expectations due to the absence of explanations, resulting in slightly higher transparency ratings. The heightened agency displayed by the robot in Learning 4 may have influenced its sociability rating, as participants may have perceived a greater degree of autonomy, potentially conflicting with the lack of responsiveness to their preferences. 

With regard to Hypothesis 2 (\textbf{H2}), our findings revealed higher transparency ratings in an approach where the robot solely considered participants' preferences. However, this approach exhibited statistical differences with the approach without preference consideration (only in Learning 4) only in the Legibility factor, with no statistical differences in the Predictability and Expectability factors. Concerning the overall Transparency score, this approach (Learning 1) considering preferences exhibited statistical differences only with the approach providing explanations without considering preferences (Learning 4). These results partially support the role of considering human preferences in enhancing transparency.

Finally, our outcomes support to Hypothesis 3 (\textbf{H3}), indicating that a robot that takes human preferences into account during the learning process and provides explanations of its actions (Learning 3) is more overall Transparent (and Expectable) than the robot employing other learning mechanisms. Participants rated Learning 3 as safer and more comfortable than the other three learning mechanisms (see Figure \ref{transp1}). Furthermore, our investigation extended to the impact of these learning mechanisms on the social attribution of the robot. The results indicate an increase in the robot's anthropomorphism when Learning 3 was employed.

In summary, our study underscores the intricate interplay of various factors influencing the perception of transparency in robot learning. The results highlight the significance of considering human preferences and providing explanations to improve transparency, as well as their effects on sociability and anthropomorphism attributions.

\subsection{Limitations}

This investigation proposes a mechanism to increase the transparency of reinforcement learning in autonomous robots. However, there are limitations that provide opportunities for future exploration.
First, more sophisticated explanations could be considered to investigate whether they impact the user's understanding of the robot's behaviours. However, the focus of this investigation was primarily on the effect of including human preferences during learning on transparency.

To further analyse the effectiveness of user preferences in driving robot decisions during learning, user modelling and profiling need to be addressed in future works.
Given the importance of personalization in Human-Robot Interaction, our findings show that it can play a crucial role when learning occurs in interactive scenarios.

Finally, it should be noted that using a shielding mechanism in Reinforcement Learning may potentially limit the agent's exploration of the environment because it restricts the agent from performing specific actions. Consequently, the agent may fail to learn an optimal policy, leading to suboptimal or incomplete learning. Furthermore, if the shielded actions are crucial for the agent to learn an effective policy, the agent's convergence to an optimal policy is not guaranteed. Nonetheless, to avoid this problem, in the present work, a more flexible shielding mechanism is proposed that always provides a feasible alternative action in case the preferable actions are not feasible. The agent can still find a policy even if it does not adhere to human preferences. 

Despite these limitations, this paper is a valuable contribution to the field, providing a roadmap for future studies. These potential improvements and directions offer opportunities to further advance our understanding of transparency in robots during the learning process.

\section{Conclusions}

This paper focuses on developing a transparent learning mechanism for robots interacting with humans. The study highlights the importance of considering human preferences/constraints on its decision-making during learning to obtain transparent robot behaviours. We propose a method for incorporating human preferences into the reinforcement learning process using a post-posed shield that might replace the robot's action with a preferable alternative based on human preferences. Four distinct learning conditions are analysed in this research, and the impact on the social attribution of the robot is assessed. 

Our results suggest that simple explanations alone might not increase transparency if the user has preferences that are not considered. However, a robot that considers people's preferences in the learning process and provides explanations of its actions increases transparency. 
The experimental results obtained by this study are encouraging, even if collected for a simple scenario. In future works, we will evaluate the robustness of the proposed approach by refining our model to include more complex human preferences that can be adopted in real-world settings. We also want to evaluate in which cases the computational cost introduced by the shielding mechanism can be balanced by a faster convergence of the RL when considering preferences.


\balance
\printbibliography

\end{document}